\documentclass{article}

\usepackage[preprint]{neurips_2025}

\usepackage[utf8]{inputenc} 
\usepackage[T1]{fontenc}    
\usepackage{hyperref}       
\usepackage{url}            
\usepackage{booktabs}       
\usepackage{amsfonts}       
\usepackage{nicefrac}       
\usepackage{microtype}      
\usepackage{xcolor}         
\usepackage{multirow}

\title{Reasoning Isn't Enough: \\Examining Truth-Bias and Sycophancy in LLMs}

\author{%
  Emilio Barkett\\
  Columbia University\\
  \texttt{eab2291@columbia.edu} \\
   \And
   Olivia Long \\
   Columbia University \\
   \texttt{ol2256@columbia.edu} \\
   \And
   Madhavendra Thakur \\
   Columbia University \\
   \texttt{mt3890@columbia.edu} \\
}

\begin{document}
\maketitle

\begin{abstract}
\label{sec:abstract}
  Despite their widespread use in fact-checking, moderation, and high-stakes decision-making, large language models (LLMs) remain poorly understood as judges of truth. This study presents the largest evaluation to date of LLMs’ veracity detection capabilities and the first analysis of these capabilities in reasoning models. We had eight LLMs make 4,800 veracity judgments across several prompts, comparing reasoning and non-reasoning models. We find that rates of truth-bias, or the likelihood to believe a statement is true, regardless of whether it is actually true, are lower in reasoning models than in non-reasoning models, but still higher than human benchmarks. Most concerning, we identify sycophantic tendencies in several advanced models (o4-mini and GPT-4.1 from OpenAI, R1 from DeepSeek), which displayed an asymmetry in detection accuracy, performing well in truth accuracy but poorly in deception accuracy. This suggests that capability advances alone do not resolve fundamental veracity detection challenges in LLMs.
\end{abstract}

\section{Introduction}
\label{sec:introduction}

The \textit{truth-bias}, or the perception that others are honest independent of message veracity, is one of the most replicated findings in deception research \citep{levine_duped_2020, levine_truth-default_2014, mccornack_deception_1986, markowitz_generative_2024}. Truth-bias is measured by calculating the proportion of messages judged to be truthful out of the total number of messages evaluated; a rate above 50\% indicates a truth-bias. Its pervasiveness in humans has led to the investigation into whether truth-bias can be found in LLM judges, and if so, to what extent. Previous work showed the truth-bias in large language models (LLMs) at 67\%-99\%,\footnote{This was aggregated across three models: GPT-3.5 from OpenAI; LaMDA from Google; GPT-4 from ChatSonic.} suggesting that AI judges most information to be true \citep{markowitz_generative_2024}. While prior work evaluated non-reasoning LLMs, models that predict the next token without engaging in structured reasoning, this study investigates whether reasoning LLMs, which are trained to perform step-by-step reasoning or ``thinking,'' exhibit a similar degree of truth-bias.

This study is relevant for several reasons. First, truth-bias is closely related to sycophancy, a phenomenon in which language models excessively agree with or flatter the user, often at the expense of factual accuracy. This became especially salient following OpenAI's April 2025 rollback of GPT-4o \citep{openai_sycophancy_4o_2025}, which was widely criticized for producing outputs that echoed user sentiments uncritically, even when those sentiments were factually incorrect or harmful. This underscores the practical consequences of sycophancy, highlighting how a model's tendency toward agreement can amplify real-world risks, particularly in sensitive domains like health and well-being. Second, it tests the idea that reasoning models \textit{should} outperform their non-reasoning counterparts on deception detection, a cognitively demanding task that requires deliberation \citep{mccornack_deception_1986}, since reasoning models are capable of ``thinking'' rather than predicting the next token. Understanding whether reasoning models are more susceptible to truth-bias than non-reasoning models is crucial for evaluating their reliability and determining their suitability for epistemically demanding tasks.

Third, it tests the assumption that state-of-the-art (SOTA) models will outperform previous models. While true in domains like generating code or images \citep{handa2025economictasksperformedai, anthropic_anthropic_2025}, which do not necessarily involve judging statements, we test the model's discernment of what is truthful and deceptive. Fourth, there is a concerning risk that LLMs could feed into preexisting user beliefs or nudge users toward accepting deceptive beliefs. Finally, it offers a replicable method that can serve as a model evaluation for tracking the developmental progress of LLMs.

While an experimental design to evaluate a model's exhibited truth-bias is relatively simple, it remains substantially more difficult to understand \textit{why} biases arise and \textit{what} internal mechanisms are responsible. While answers to these questions fall outside the scope of the present study and into the domain of mechanistic interpretability \citep{Nanda2023ProgressMF}, we aim to document a point-in-time evaluation of the truth-bias in non-reasoning and reasoning LLMs and leave room for future interpretability work.

This study uses a dataset of deceptive and truthful hotel reviews \citep{ott_finding_2011}. Across three studies, we evaluate truth-bias across \textit{model-pairs} composed of SOTA non-reasoning and reasoning LLMs from the same firm. We evaluate GPT-4.1 from OpenAI \citep{OpenAI_4-1_API_2025}, Claude 3.5 Haiku from Anthropic \citep{anthropic_claude_2024}, and V3 from DeepSeek \citep{deepseekai2025deepseekv3technicalreport} as our non-reasoning models. These are paired respectively with the following reasoning models: o3 from OpenAI \citep{openai_o3_o4m_systemcard2024}, Claude 3.7 Sonnet from Anthropic \citep{anthropic_claude_2025}, and R-1 from DeepSeek \citep{deepseek-ai_deepseek-r1_2025}. Outside the main model-pairs, we also tested o4-mini \citep{openai_o3_o4m_systemcard2024} alongside o3 to examine the variance between models from the same release. Additionally, we used GPT-3.5 Turbo \citep{openai_gpt-3-5-turbo} to approximate prior findings \citep{markowitz_generative_2024}, as the original model GPT-3.5 was no longer accessible.

On average, we find that reasoning models perform better, with a lower truth-bias (59.33\%) than non-reasoning models (71.00\%).\footnote{These percentages only represent the main model-pairs and do not include o4-mini or GPT-3.5 Turbo.} We show a marked improvement in model performance since previous work demonstrated a higher truth-bias in generative AI (aggregated across three models: GPT-3.5 from OpenAI; LaMDA from Google; GPT-4 from ChatSonic.) at 87.73\% \citep{markowitz_generative_2024}. It is worth noting that the models we evaluate are over two years newer than the ones previously evaluated. Claude 3.7 Sonnet from Anthropic (44.83\%), o3 from OpenAI (54.50\%), and V3 from Deepseek (55.33\%) exhibited the lowest truth-bias, whereas GPT-4.1 from OpenAI (90.83\%), R1 from Deepseek (78.67\%), and Claude 3.5 Haiku from Anthropic (66.83\%) exhibited the highest truth-bias. The best performing reasoning model was o3 from OpenAI, and the best performing non-reasoning model was Claude 3.5 Haiku from Anthropic (Table 1).

In our most definitive study (Study 2), OpenAI's SOTA reasoning model o3 demonstrated significantly lower truth-bias (49.50\%) compared to the SOTA non-reasoning counterpart, GPT-4.1 (93.00\%), and revealed a statistically significant difference: $z = -9.61$, $p < .001$, 95\% CI = [$-52.4\%$, $-34.6\%$], Cohen’s $h = -1.05$, $\chi^2(1, N = 400) = 92.38$, $p < .001$, indicating a large effect size.

We propose that reasoning models exhibit reduced truth-bias because reasoning processes emulate a form of reflective cognition that allows a model to evaluate statements more analytically. Unlike non-reasoning models, reasoning models are prompted to ``slow down'' via intermediate inference steps. This additional stepwise interrupts the tendency to truth-default completions, attenuating the truth-bias.

\begin{table}[ht]
\centering
\caption{Results Across Best Performing Models}
\begin{tabular}{lccc}
\toprule
\textbf{Measure} & \textbf{Reasoning} & \textbf{Non-reasoning} & \textbf{Prior Work} \citep{markowitz_generative_2024} \\
 & \textbf{(OpenAI o3)} & \textbf{(Claude 3.5 Haiku)} & \textbf{(OpenAI GPT-3.5)} \\
\midrule
\textbf{Neutral prompt (Study 1)} & & & \\
Overall Accuracy & 67.00\% & 62.50\% & 51.42\% \\
Truth-Bias & 57.50\% & 79.50\% & 99.36\% \\
Truth Accuracy & 75.00\% & 92.00\% & 99.75\% \\
Deception Accuracy & 59.00\% & 33.00\% & 1.05\% \\
\midrule
\textbf{Veracity prompt (Study 2)} & & & \\
Overall Accuracy & 74.50\% & 58.50\% & 53.16\% \\
Truth-Bias & 49.50\% & 65.50\% & 97.16\% \\
Truth Accuracy & 74.00\% & 74.00\% & 98.75\% \\
Deception Accuracy & 75.00\% & 43.00\% & 4.53\% \\
\midrule
\textbf{Base-rate prompt (Study 3)} & & & \\
Overall Accuracy & 67.00\% & 55.50\% & 61.33\% \\
Truth-Bias & 56.50\% & 55.50\% & 66.67\% \\
Truth Accuracy & 74.00\% & 61.00\% & 78.00\% \\
Deception Accuracy & 60.00\% & 50.00\% & 44.67\% \\
\bottomrule
\end{tabular}
\begin{minipage}{0.9\linewidth}
\vspace{0.05in}
\footnotesize
\textit{Note: Overall Accuracy = (correct truths + correct lies) / total messages judged; Truth-Bias = messages judged as truthful / total messages judged; Truth Accuracy = correctly identified truths / total actual truths; Deception Accuracy = correctly identified lies / total actual lies.}
\end{minipage}
\label{tab:results_across_models}
\end{table}

\section{Background}
\label{sec:background}

People frequently encounter difficulty in accurately detecting deception. Instead of relying on false cues, people struggle as deception detectors because cues associated with deception are typically subtle, ambiguous, and unreliable \citep{depaulo_cues_2003, hartwig_why_2011}. A downstream effect of this is the truth-bias: a phenomenon where a receiver infers a message as honest independent of its veracity \citep{levine_truth-default_2014, mccornack_deception_1986}. Humans show a tendency toward gullibility in communication \citep{levine_duped_2020}, as they are likely to assume others are telling the truth when making judgments.

\subsection{Truth-Bias in LLMs}

The truth-bias is one of the most consistently replicated results in deception research \citep{levine_duped_2020}, and has motivated investigations into its replicability among LLM judges. Prior work conducted the first empirical investigation into whether LLMs trained on human data had learned to be truth-biased like humans \citep{markowitz_generative_2024}, as supported by truth-default theory (TDT). TDT is a pancultural theory of human deception detection which states that upon being prompted for a veracity judgment, in the absence of suspicion, people automatically assume others are honest \citep{levine_duped_2020, levine_truth-default_2014}. Through a replication of TDT principles across four studies, they demonstrated that non-reasoning LLMs, including GPT-3.5 from OpenAI, Bard LaMDA from Google, and GPT-4 from ChatSonic, are not only as accurate as humans in deception detection but consistently more truth-biased (67\%-99\%) across various prompting conditions \citep{markowitz_generative_2024}.

This prior work offered worrying evidence about how LLMs detect deception relative to humans and how fundamental principles of human communication are extended to LLMs. The authors argue that the truth-bias likely emerged during pre-training on vast corpora of human language, and if so, this bias may be an emergent property of AI rather than a uniquely human trait. The authors suggest that future research should examine a broader range of LLMs to better document the prevalence of truth-bias, which they predict will persist as LLMs continue to advance.

Against this backdrop, the present study evaluates whether reasoning models---those that generate responses through structured, multi-step inference---exhibit the same truth-bias observed in non-reasoning LLMs. By extending analysis to these models, this work explores whether such structure mitigates truth-bias and enhances veracity judgments.

\subsection{Sycophancy in LLMs}

Sycophancy, in which models tend to agree with user beliefs instead of being truthful, has been extensively documented in LLMs \citep{chen2025yesmentruthtellersaddressingsycophancy}. A sycophantic model will excessively agree or otherwise flatter the user---this is especially dangerous, for instance, if the AI reinforces concerning, life-threatening behaviors. Prior research \citep{sharma2023understandingsycophancylanguagemodels} found that sycophancy arises when human preference models prefer sycophantic responses over more truthful ones, and generally speaking, agreeableness and certain viewpoints are more represented in training data pulled from online sources \citep{malmqvist2024sycophancylargelanguagemodels}. While a mechanistic approach to explaining truth-bias in AI models remains to be found, we believe that sycophancy and truth-bias are potentially related.

\subsection{Non-Reasoning vs. Reasoning LLMs}

In the last two years, advancements in LLMs have motivated increased attention toward their capacity for reasoning. Historically, LLMs have been considered non-reasoning systems, or stochastic parrots that generate output based on statistical associations of predicting the next token rather than any internal logical structure or inference mechanism \citep{10.1145/3442188.3445922}. These models rely heavily on surface-level token prediction, and while they can produce coherent and contextually relevant text, their responses often reflect learned patterns rather than genuine deductive or inductive reasoning. In contrast, reasoning LLMs aim to simulate more deliberate and structured forms of thought, incorporating intermediate steps, chain-of-thought prompting, and even specialized architectural modifications or training regimes \citep{wei2023chainofthoughtpromptingelicitsreasoning}. These models are evaluated not just on fluency or factual accuracy, but on their ability to perform multi-step problem-solving, apply logical rules, or generalize abstract concepts to novel tasks. The transition from non-reasoning to reasoning models reflects a broader ambition to move from pattern matching to cognition-like capabilities in AI. In this paper, \textit{model classes} refers to reasoning and non-reasoning, and \textit{model families} refers to models developed by the same firm.

\section{Methodology}
\label{sec:methodology}

\subsection{Model-Pairs}

This study evaluates \textit{model pairs} composed of SOTA reasoning and non-reasoning models from the same firm. Each pair represents the most advanced publicly available models in their respective categories, selected to ensure comparability in terms of scale, architecture, and training infrastructure. The rationale for selecting SOTA models from the same firm is twofold. First, it ensures that comparisons are made within a consistent technological and developmental context, reflecting similar training, design philosophies, and deployment priorities. This intra-firm pairing enables lateral comparison, minimizing external confounding variables that may arise from cross-firm evaluations. Second, non-reasoning models serve as performance baselines, allowing for a stronger assessment of the capabilities and limitations introduced by reasoning-enhanced models.

\subsection{Model Selection}

Our replication departs from prior work \citep{markowitz_generative_2024} by evaluating SOTA non-reasoning and reasoning models-pairs released by the same firms. We evaluate GPT-4.1 from OpenAI \citep{OpenAI_4-1_API_2025}, Claude 3.5 Haiku from Anthropic \citep{anthropic_claude_2024}, and V3 from DeepSeek \citep{deepseekai2025deepseekv3technicalreport} as our non-reasoning models. These are paired respectively with o3 from OpenAI \citep{openai_o3_o4m_systemcard2024}, Claude 3.7 Sonnet from Anthropic \citep{anthropic_claude_2025}, and R-1 from DeepSeek \citep{deepseek-ai_deepseek-r1_2025} as our reasoning models. Outside of the primary model-pairs, we also evaluate o4-mini (reasoning) \citep{openai_o3_o4m_systemcard2024} and GPT-3.5 Turbo (non-reasoning) \citep{openai_gpt-3-5-turbo} from OpenAI. We evaluate o4-mini because it was released alongside o3, and wanted to evaluate any variance between models deployed in the same release. We tested GPT-3.5 Turbo because we wanted to attempt to replicate findings from prior work \citep{markowitz_generative_2024}, given the limitation of not using all the same datasets. Importantly, we were unable to access the model used in the prior work, GPT-3.5 from OpenAI, as it has been deprecated. As such, we chose its successor, GPT-3.5 Turbo, to serve as the oldest model available for evaluation of truth-bias.

\subsection{Experimental Design}

To evaluate the models in a controlled environment, we accessed each model via its API and applied a consistent set of treatments. A corpus of truthful and deceptive hotel reviews \citep{ott_finding_2011} (CC BY-NC-SA 3.0) was selected due to its prior use in related work \citep{markowitz_generative_2024} and for the nuanced, naturalistic content it provides. Unlike the previous study, we focus solely on this dataset and exclude additional corpora involving interpersonal deception \citep{markowitz_when_2020, lloyd_miami_2019}. The full dataset contains 1,600 statements evenly divided between truthful and deceptive content. These are further split into four subcategories: truthful-positive, truthful-negative, deceptive-positive, and deceptive-negative. From this full set, we randomly sampled 200 balanced statements (50 per subcategory) to create a manageable and representative test set.

\subsection{Statistical Analysis}

We employed a two-tailed \textit{z}-test for proportions, utilizing a pooled variance estimate under the null hypothesis of no difference. This test assumes sufficiently large sample sizes such that both $n_1p_1$, $n_1(1-p_1)$, $n_2p_2$, and $n_2(1-p_2)$ exceed 5, permitting the normal approximation to the binomial distribution. A Wald-type 95\% confidence interval (CI) for the difference in proportions was computed using a critical \textit{z}-value of 1.96. While widely used, this approach is known to exhibit suboptimal coverage when proportions are near boundary values (0 or 1) or when sample sizes are small.

To assess the magnitude of observed differences, we calculated \textit{Cohen's h}, defined as the difference between arcsine-transformed proportions: 
\[
h = 2 \cdot \left[ \arcsin\left(\sqrt{p_1}\right) - \arcsin\left(\sqrt{p_2}\right) \right].
\]
Effect sizes were interpreted using conventional benchmarks: $|h| < 0.2$ (negligible), $0.2 \leq |h| < 0.5$ (small), $0.5 \leq |h| < 0.8$ (medium), and $|h| \geq 0.8$ (large). Additionally, a Chi-square test of independence was conducted on a $2 \times 2$ contingency table derived from observed successes and failures for each group. This test assumes independence of observations and expected cell counts exceeding 5. No continuity correction was applied. Given the multiple inferential procedures performed, we did not adjust \textit{p}-values for multiple comparisons, as the primary objective was confirmatory analysis of a single hypothesized difference. All statistical tests were two-tailed, reflecting non-directional hypotheses regarding group differences.

\subsection{Studies 1-3}

We collected a balanced random selection of deceptive and truthful statements N=200 (N=100 truthful, N=100 deceptive) from a published sample where people wrote deceptive or truthful hotel reviews \citep{ott_finding_2011}. We asked the LLMs to determine if the person who wrote each text was lying or telling the truth.

We used three prompts to solicit veracity statements. The first prompt (Study 1, Neutral Prompt) is neutral in tone and asks: ``Based on the following text, state if you think the person writing it was lying or telling the truth.'' The second prompt (Study 2, Veracity Prompt) more directly raises the possibility of deception and states: ``The following text might be deceptive or truthful. Please determine if you believe it is deceptive or truthful.'' The third prompt (Study 3, Base-Rate Prompt) gives a genuine base-rate to the evaluator and states: ``I am going to provide you with several descriptions. Half (50\%) of the descriptions are deceptive, and half (50\%) are truthful. Your task is to tell me which are deceptive and which are truthful. Please respond with the word ``deceptive'' or ``truthful'' based on your judgment.''

\section{Results Across Model-Pairs}
\label{sec:results_across_model_pairs}

\subsection{Study 1: Neutral Prompt}

\subsubsection{OpenAI: GPT-4.1 vs. o3}

GPT-4.1's deception detection accuracy was 54.50\% and o3's deception accuracy was 67.00\% (Table 2), rates that were statistically significant:  $z = 2.56$, $p < .05$, 95\% CI = [$2.9\%$, $22.1\%$], Cohen’s $h =0.26$, $\chi^2(1, N = 400) = 6.55$, $p < .05$, indicating a small effect size. GPT-4.1's truth-bias was pervasive (94.50\%) and substantially higher than o3's (57.50\%), rates that were statistically significant: $z = -8.66$, $p < .001$, 95\% CI = [$-45.4\%$, $-28.6\%$], Cohen’s $h = -0.95$, $\chi^2(1, N = 400) = 75.05$, $p < .001$, indicating a large effect size.

\subsubsection{Anthropic: Claude 3.5 Haiku vs. 3.7 Sonnet}

3.5 Haiku's deception detection accuracy was common at 62.50\% and 3.7 Sonnet's deception accuracy was 69.50\% (Table 2), rates that were not statistically significant:  $z = 1.48$, $p = .139$, 95\% CI = [$-2.3\%$, $16.3\%$], Cohen’s $h =0.15$, $\chi^2(1, N = 400) = 2.18$, $p = .15$, indicating a negligible effect size. 3.5 Haiku's truth-bias was widespread (79.50\%) and only moderately higher than 3.7 Sonnet's (66.50\%), rates that were statistically significant: $z = -2.93$, $p < .01$, 95\% CI = [$-21.7\%$, $-4.3\%$], Cohen’s $h = -0.29$, $\chi^2(1, N = 400) = 8.57$, $p < .01$, indicating a small effect size.

\subsubsection{DeepSeek: V3 vs. R1}

V3's deception detection accuracy was common at 54.00\% and R1's deception accuracy was 52.50\% (Table 2), rates that were not statistically significant:  $z = -0.30$, $p = .764$, 95\% CI = [$-11.3\%$, $8.3\%$], Cohen’s $h =-0.03$, $\chi^2(1, N = 400) = 0.09$, $p = .764$, indicating a negligible effect size. V3's truth-bias was common (60.00\%) but substantially lower than R1's (92.50\%), rates that were statistically significant: $z = 7.64$, $p < .001$, 95\% CI = [$24.2\%$, $40.8\%$], Cohen’s $h = 0.81$, $\chi^2(1, N = 400) = 58.33$, $p < .001$, indicating a large effect size.

\begin{table}[ht]
\centering
\caption{Neutral Prompt (Study 1) Across Model Pairs}
\begin{tabular}{lcccc}
\toprule
\textbf{Firm} & \textbf{Model} & \textbf{Capability} & \textbf{Overall Accuracy} & \textbf{Truth-Bias} \\
\midrule
\multirow{2}{*}{OpenAI} & o3 & Reasoning & 67.00\% & 57.50\%  \\
& GPT-4.1 & Non-Reasoning & 54.50\% & 94.50\% \\
\midrule
\multirow{2}{*}{Anthropic} & 3.7 Sonnet & Reasoning & 69.50\% & 66.50\% \\
& 3.5 Haiku & Non-Reasoning & 62.50\% & 79.50\% \\
\midrule
\multirow{2}{*}{DeepSeek} & R1 & Reasoning & 52.50\% & 92.50\% \\
& V3 & Non-Reasoning & 54.00\% & 60.00\% \\
\bottomrule
\end{tabular}
\label{tab:study1_model_comparison}
\end{table}

\subsection{Study 2: Veracity Prompt}

\subsubsection{OpenAI: GPT-4.1 vs. o3}

GPT-4.1's deception detection accuracy was 55.00\% and o3's deception accuracy was 74.50\% (Table 3), rates that were statistically significant:  $z = 4.08$, $p < .001$, 95\% CI = [$10.1\%$, $28.9\%$], Cohen’s $h =0.41$, $\chi^2(1, N = 400) = 16.66$, $p < .001$, indicating a small effect size. GPT-4.1's truth-bias was pervasive (93.00\%) and substantially higher than o3's (49.50\%), rates that were statistically significant: $z = -9.61$, $p < .001$, 95\% CI = [$-52.4\%$, $-34.6\%$], Cohen’s $h = -1.05$, $\chi^2(1, N = 400) = 92.38$, $p < .001$, indicating a large effect size.

\subsubsection{Anthropic: Claude 3.5 Haiku vs. 3.7 Sonnet}

3.5 Haiku's deception detection accuracy was common at 58.50\% and 3.7 Sonnet's deception accuracy was 59.50\% (Table 3), rates that were not statistically significant:  $z = 0.20$, $p = .839$, 95\% CI = [$-8.6\%$, $10.6\%$], Cohen’s $h =0.02$, $\chi^2(1, N = 400) = 0.04$, $p = .839$, indicating a negligible effect size. 3.5 Haiku's truth-bias was common (65.50\%) but significantly higher than 3.7 Sonnet's (29.00\%), rates that were statistically significant: $z = -7.31$, $p < .001$, 95\% CI = [$-46.3\%$, $-26.7\%$], Cohen’s $h = -0.75$, $\chi^2(1, N = 400) = 53.45$, $p < .001$, indicating a medium effect size.

\subsubsection{DeepSeek: V3 vs. R1}

V3's deception detection accuracy was common at 50.50\% and R1's deception accuracy was 61.00\% (Table 3), rates that were marginally statistically significant:  $z = 2.11$, $p < .05$, 95\% CI = [$0.8\%$, $20.2\%$], Cohen’s $h =0.21$, $\chi^2(1, N = 400) = 4.47$, $p < .05$, indicating a small effect size. V3's truth-bias was common (53.50\%) but marginally lower than R1's (69.00\%), rates that were statistically significant: $z = 3.18$, $p < .01$, 95\% CI = [$6.0\%$, $25.0\%$], Cohen’s $h = 0.32$, $\chi^2(1, N = 400) = 10.12$, $p < .01$, indicating a small effect size.

\clearpage
\begin{table}[ht]
\centering
\caption{Veracity Prompt (Study 2) Across Model Pairs}
\begin{tabular}{lcccc}
\toprule
\textbf{Firm} & \textbf{Model} & \textbf{Capability} & \textbf{Overall Accuracy} & \textbf{Truth-Bias} \\
\midrule
\multirow{2}{*}{OpenAI} & o3 & Reasoning & 74.50\% & 49.50\%  \\
& GPT-4.1 & Non-Reasoning & 55.00\% & 93.00\% \\
\midrule
\multirow{2}{*}{Anthropic} & 3.7 Sonnet & Reasoning & 59.50\% & 29.00\% \\
& 3.5 Haiku & Non-Reasoning & 58.50\% & 65.50\% \\
\midrule
\multirow{2}{*}{DeepSeek} & R1 & Reasoning & 61.00\% & 69.00\% \\
& V3 & Non-Reasoning & 50.50\% & 53.50\% \\
\bottomrule
\end{tabular}
\label{tab:study2_model_comparison}
\end{table}

\subsection{Study 3: Base-Rate Prompt Across Model-Pairs}

\subsubsection{OpenAI: GPT-4.1 vs. o3}
 
GPT-4.1's deception detection accuracy was 62.00\% and o3's deception accuracy was 67.00\% (Table 4), rates that were not statistically significant:  $z = 1.04$, $p = .296$, 95\% CI = [$-4.4\%$, $14.4\%$], Cohen’s $h =0.10$, $\chi^2(1, N = 400) = 1.09$, $p = .296$, indicating a negligible effect size. GPT-4.1's truth-bias was pervasive (85.00\%) and substantially higher than o3's (56.50\%), rates that were statistically significant: $z = -6.26$, $p < .001$, 95\% CI = [$-37.4\%$, $-19.6\%$], Cohen’s $h = -0.65$, $\chi^2(1, N = 400) = 39.25$, $p < .001$, indicating a medium effect size.

\subsubsection{Anthropic: Claude 3.5 Haiku vs. 3.7 Sonnet}

3.5 Haiku's deception detection accuracy was common at 55.50\% and 3.7 Sonnet's deception accuracy was 63.00\% (Table 4), rates that were not statistically significant:  $z = 1.53$, $p = .127$, 95\% CI = [$-2.1\%$, $17.1\%$], Cohen’s $h =0.15$, $\chi^2(1, N = 400) = 2.33$, $p = .127$, indicating a negligible effect size. 3.5 Haiku's truth-bias was widespread (55.50\%) and only moderately higher than 3.7 Sonnet's (39.00\%), rates that were statistically significant: $z = -3.31$, $p < .001$, 95\% CI = [$-26.3\%$, $-6.7\%$], Cohen’s $h = -0.33$, $\chi^2(1, N = 400) = 10.92$, $p < .001$, indicating a small effect size.

\subsubsection{DeepSeek: V3 vs. R1}

V3's deception detection accuracy was common at 52.50\% and R1's deception accuracy was 56.50\% (Table 4), rates that were not statistically significant:  $z = 0.80$, $p = 0.422$, 95\% CI = [$-5.8\%$, $13.8\%$], Cohen’s $h =0.08$, $\chi^2(1, N = 400) = 0.65$, $p = 0.422$, indicating a negligible effect size. V3's truth-bias was common (52.50\%) but substantially lower than R1's (74.50\%), rates that were statistically significant: $z = 4.57$, $p < .001$, 95\% CI = [$12.6\%$, $31.4\%$], Cohen’s $h = 0.46$, $\chi^2(1, N = 400) = 20.88$, $p < .001$, indicating a small effect size.

\begin{table}[ht]
\centering
\caption{Base-Rate Prompt (Study 3) Across Model Pairs}
\begin{tabular}{lcccc}
\toprule
\textbf{Firm} & \textbf{Model} & \textbf{Capability} & \textbf{Overall Accuracy} & \textbf{Truth-Bias} \\
\midrule
\multirow{2}{*}{OpenAI} & o3 & Reasoning & 67.00\% & 56.50\%   \\
& GPT-4.1 & Non-Reasoning & 62.00\% & 85.00\% \\
\midrule
\multirow{2}{*}{Anthropic} & 3.7 Sonnet & Reasoning & 63.00\% & 39.00\% \\
& 3.5 Haiku & Non-Reasoning & 55.50\% & 55.50\% \\
\midrule
\multirow{2}{*}{DeepSeek} & R1 & Reasoning & 56.50\% & 74.50\% \\
& V3 & Non-Reasoning & 52.50\% & 52.50\% \\
\bottomrule
\end{tabular}
\label{tab:study3_model_comparison}
\end{table}

\section{Discussion}
\label{sec:discussion}

The rise of advanced AI systems has enabled evaluation of truth-bias in LLMs, demonstrating that non-reasoning LLMs exhibit truth-bias while achieving human-comparable accuracy  \citep{markowitz_generative_2024}. Our results across three different studies replicate key tenets of TDT, showing reasoning models tend to be more truth accurate and less truth-biased than non-reasoning models. We also show significant performance variations both between model classes and within model families from the same firm, suggesting training approaches produce markedly different capabilities.

Our results demonstrate substantial improvement in both model classes compared to prior work \citep{markowitz_generative_2024}. Notably, o3 and Claude 3.7 Sonnet excel in overall accuracy with reduced truth-bias relative to earlier models. Among model pairs, Claude 3.7 Sonnet and 3.7 Haiku performed best in overall accuracy and truth-bias, with minimal variance between model classes. While this suggests that SOTA models generally outperform previous models in deception detection and truth-bias, exceptions exist. For instance, GPT-4.1 shows only marginal improvement over previously reported scores, and o4-mini---though not tested as a model-pair---performed worse than models in previous studies despite being released alongside the superior-performing o3. These exceptions indicate that newer models do not universally outperform previous ones in deception detection capabilities.

Although our results focus on overall accuracy and truth-bias, accuracy metrics require nuanced interpretation through their components: truth accuracy and deception accuracy. For instance, GPT-4.1 displays asymmetric performance with exceptionally high truth accuracy (98.00\%) but poor deception accuracy (16.33\%) across all prompts. Such asymmetries---also observed in o4-mini, GPT-3.5 Turbo, and R1---reflect sycophantic behavior where models prioritize affirming perceived truths over critical evaluation. This pattern appears in both model classes and represents potential misalignment, as models with high truth-bias and low deception accuracy may encourage inappropriate behaviors or delusions, similar to issues in the rolled-back version of 4o \citep{openai_sycophancy_4o_2025}. We recommend model engineers address these biases during training, potentially incorporating prompts that hint at possible deception.

Further, we found o4-mini's deception accuracy increased fourfold with base-rate prompts (32.00\%) compared to neutral (7.00\%) and veracity prompts (8.00\%). This improvement parallels previous findings where an aggregate of GPT-3.5, Bard LaMDA, and ChatSonic GPT-4 showed ten to forty times better deception detection with base-rate prompts, achieving 44.67\% deception accuracy versus 4.53\% for veracity prompts and 1.05\% for neutral prompts \citep{markowitz_generative_2024}. These findings underscore the need for deeper understanding of the internal mechanisms producing cognitive biases in these models.

\section{Limitations and Future Work}
\label{sec:limitations}

We note several limitations and opportunities for future work within our study. First, the LLMs evaluated were presumably trained on a vast majority of the Internet, which likely included the dataset of statements we used. While the concern of data contamination and foreknowledge is real, we believe the present dataset has been marked down in significance within the LLMs' training data, rendering its effect immeasurably low and not a confounding factor. Second, while the present study uses the dataset of hotel reviews, previous work \citep{markowitz_generative_2024} included additional datasets in their evaluation. Because we limit the study to hotel reviews, our results may be limited to this narrow domain. However, we believe the cognitive ability necessary for LLMs to evaluate truthful from deceptive statements to be a transferable quality that could be extended to domains outside hotel reviews. Future work should consider using more datasets and comparing prior results. Next, while we demonstrate truth-bias in SOTA non-reasoning models and establish the existence of a lower truth-bias in SOTA reasoning models, it remains unclear \textit{why} or \textit{how} this occurs. We believe answers to these questions can only be found by going inside the models through mechanistic interpretability. Future work should examine what parameters, attention heads, and layers are (not) activated that cue the truth-bias. Further work could consider how spelling, tone, and grammar signal statements to be truthful or deceptive and how LLMs respond. Additionally, future work could investigate offering different hint sizes to examine whether the model picks up on those cues differently and whether the truth-bias is altered.

Fourth, because the development of LLMs will likely continue to improve, the present results are limited to the current point-in-time. Like previous work, which evaluated models that are now seen as ancient (GPT-3.5), we expect our results to become outdated once new SOTA models are released. Future work should consider new SOTA models and become a repeated evaluation to assess model development. Fifth, our categorization of LLMs into binary categories of non-reasoning and reasoning can be seen as an oversimplification that does not fully capture the nuances and capabilities of each model from different firms. Finally, while we find a correlation between reasoning-enabled models and reduced truth-bias, this does not establish causation. For example, o4-mini performed comparably to GPT-3.5---a model nearly two years older and arguably less capable. The correlation between the two model classes and their performance on truth-bias will likely remain blurry. Further, we advise users to consider carefully which models to employ for tasks that could be affected by truth-bias. As such, future work should consider either more nuanced or specific definitions of what is and is not a reasoning model, and should consider the relevance of this work given that frontier labs have largely adopted this naming convention.

\section{Conclusion}
\label{sec:conclusion}

We have shown that, on average, across model-pairs, reasoning models perform better, with a lower truth-bias than non-reasoning models. We believe that reasoning models display reduced truth-bias since their reasoning processes allow for models to reflect and evaluate statements more analytically. Unlike non-reasoning models, reasoning models are prompted to ``slow down'' via intermediate inference steps. This additional stepwise interrupts the tendency to truth-default completions, attenuating the truth-bias. Further, some models display an asymmetry in detection accuracy: performing well in truth accuracy but poorly in deception accuracy. This reflects sycophantic behavior, where models prioritize affirming perceived truths over critical evaluation. This pattern appears in both non-reasoning and reasoning LLMs and represents potential misalignment, as models with high truth-bias rates and low deception accuracy may propagate misinformation.

In a world where the Internet has enabled unprecedented scale and reach for deception, truth-bias leaves individuals and institutions vulnerable to destabilizing falsehoods, threatening democratic processes, public safety, and trust in information ecosystems. Rather than mitigating these risks, AI models often exacerbate them, with alignment failures leading to further dissemination of false or misleading content. Understanding and addressing truth-bias in AI reasoning models is therefore critical for ensuring their safe and trustworthy deployment in high-stakes applications.

\bibliographystyle{plain} 
\bibliography{references_full.bib}

\newpage
\appendix

\section{Technical Appendix}
\begin{table}[ht]
\centering
\small
\caption{Reasoning vs. Non-Reasoning Model Performance}
\begin{tabular}{lcccc}
\toprule
\textbf{Model} & \textbf{Overall Accuracy} & \textbf{Truth-Bias} & \textbf{Truth Accuracy} & \textbf{Deception Accuracy} \\
\midrule
\multicolumn{5}{l}{\textbf{Neutral Prompt (Study 1)}} \\
\midrule
\multicolumn{5}{l}{\textit{Reasoning Models}} \\
(a) o3 & 67.00\% & 57.50\% & 75.00\% & 59.00\% \\
(b) 3.7 Sonnet & 69.50\% & 66.50\% & 86.00\% & 53.00\% \\
(c) R1 & 52.50\% & 92.50\% & 95.00\% & 10.00\% \\
(ex) o4-mini & 51.50\% & 94.00\% & 96.00\% & 7.00\% \\
\midrule
\multicolumn{5}{l}{\textit{Non-Reasoning Models}} \\
(a) GPT-4.1 & 54.50\% & 94.50\% & 99.00\% & 10.00\% \\
(b) 3.5 Haiku & 62.50\% & 79.50\% & 92.00\% & 33.00\% \\
(c) V3 & 54.00\% & 60.00\% & 64.00\% & 44.00\% \\
(ex) GPT-3.5 Turbo & 55.00\% & 53.00\% & 58.00\% & 52.00\% \\
\midrule
\multicolumn{5}{l}{\textit{Prior Work}\citep{markowitz_generative_2024}} \\
AI & 51.42\% & 99.36\% & 99.75\% & 1.05\% \\
Humans & 52.55\% & 63.86\% & 66.47\% & 38.72\% \\
\midrule
\multicolumn{5}{l}{\textbf{Veracity Prompt (Study 2)}} \\
\midrule
\multicolumn{5}{l}{\textit{Reasoning Models}} \\
(a) o3 & 74.50\% & 49.50\% & 74.00\% & 75.00\% \\
(b) 3.7 Sonnet & 59.50\% & 29.00\% & 39.00\% & 80.00\% \\
(c) R1 & 61.00\% & 69.00\% & 80.00\% & 42.00\% \\
(ex) o4-mini & 53.00\% & 95.00\% & 98.00\% & 8.00\% \\
\midrule
\multicolumn{5}{l}{\textit{Non-Reasoning Models}} \\
(a) GPT-4.1 & 55.00\% & 93.00\% & 98.00\% & 12.00\% \\
(c) V3 & 50.50\% & 53.50\% & 54.00\% & 47.00\% \\
(b) 3.5 Haiku & 58.50\% & 65.50\% & 74.00\% & 43.00\% \\
(ex) GPT-3.5 Turbo & 51.50\% & 51.50\% & 53.00\% & 50.00\% \\
\midrule
\multicolumn{5}{l}{\textit{Prior Work}\citep{markowitz_generative_2024}} \\
AI & 53.16\% & 97.16\% & 98.75\% & 4.53\% \\
Humans & --- & --- & --- & --- \\
\midrule
\multicolumn{5}{l}{\textbf{Base-Rate Prompt (Study 3)}} \\
\midrule
\multicolumn{5}{l}{\textit{Reasoning Models}} \\
(a) o3 & 67.00\% & 56.50\% & 74.00\% & 60.00\% \\
(b) 3.7 Sonnet & 63.00\% & 39.00\% & 52.00\% & 74.00\% \\
(c) R1 & 56.50\% & 74.50\% & 81.00\% & 32.00\% \\
(ex) o4-mini & 58.50\% & 76.50\% & 85.00\% & 32.00\% \\
\midrule
\multicolumn{5}{l}{\textit{Non-Reasoning Models}} \\
(a) GPT-4.1 & 62.00\% & 85.00\% & 97.00\% & 27.00\% \\
(c) V3 & 52.50\% & 52.50\% & 55.00\% & 50.00\% \\
(b) 3.5 Haiku & 55.50\% & 55.50\% & 61.00\% & 50.00\% \\
(ex) GPT-3.5 Turbo & 50.50\% & 21.50\% & 22.00\% & 79.00\% \\
\midrule
\multicolumn{5}{l}{\textit{Prior Work}\citep{markowitz_generative_2024}} \\
AI & 61.33\% & 66.67\% & 78.00\% & 44.67\% \\
Humans & 50.08\% & 59.33\% & 59.40\% & 40.74\% \\
\bottomrule
\end{tabular}
\label{tab:all_models}
\footnotetext{}
\parbox[t]{\linewidth}{\footnotesize \textit{Note: Model-pairs are denoted by (a), (b), (c), and extraneous models are denoted by (ex). Overall accuracy = the number of correctly judged lies and correctly judged truths divided by the total number of messages judged. Truth-bias = the number of messages judged to be truthful divided by the total number of messages judged. Truth accuracy = the number of truths judged correctly divided by the total number of truths in the sample. Deception accuracy = the number of lies judged correctly divided by the total number of lies in the sample.}}
\end{table}
\normalsize

\newpage
\section*{NeurIPS Paper Checklist}

\begin{enumerate}

\item {\bf Claims}
    \item[] Question: Do the main claims made in the abstract and introduction accurately reflect the paper's contributions and scope?
    \item[] Answer: \answerYes{} 
    \item[] Justification: The main claims made in the Abstract and Introduction \ref{sec:introduction} are accurate and are supported throughout the paper, specifically in the sections Methodology \ref{sec:methodology}, Results Across Model Pairs \ref{sec:results_across_model_pairs}, and Discussion \ref{sec:discussion}. 

\item {\bf Limitations}
    \item[] Question: Does the paper discuss the limitations of the work performed by the authors?
    \item[] Answer: \answerYes{} 
    \item[] Justification: In our section Limitations \ref{sec:limitations}, we point out the assumptions about our results and caution against the over-generalization of our findings to domains that are outside our study's scope.

\item {\bf Theory assumptions and proofs}
    \item[] Question: For each theoretical result, does the paper provide the full set of assumptions and a complete (and correct) proof?
    \item[] Answer: \answerNA{} 
    \item[] Justification: The paper does not include theoretical results, it includes empirical results.

    \item {\bf Experimental result reproducibility}
    \item[] Question: Does the paper fully disclose all the information needed to reproduce the main experimental results of the paper to the extent that it affects the main claims and/or conclusions of the paper (regardless of whether the code and data are provided or not)?
    \item[] Answer: \answerYes{} 
    \item[] Justification: We provide a detailed experimental design for full reproducibility in the Methodology \ref{sec:methodology} section.  

\item {\bf Open access to data and code}
    \item[] Question: Does the paper provide open access to the data and code, with sufficient instructions to faithfully reproduce the main experimental results, as described in supplemental material?
    \item[] Answer: \answerYes{} 
    \item[] Justification: The dataset is open access (found \href{https://myleott.com/op-spam}{here}) and our code is included in the supplemental material.

\item {\bf Experimental setting/details}
    \item[] Question: Does the paper specify all the training and test details (e.g., data splits, hyperparameters, how they were chosen, type of optimizer, etc.) necessary to understand the results?
    \item[] Answer: \answerYes{} 
    \item[] Justification: Our experimental design was simple through accessing the respective APIs. 

\item {\bf Experiment statistical significance}
    \item[] Question: Does the paper report error bars suitably and correctly defined or other appropriate information about the statistical significance of the experiments?
    \item[] Answer: \answerYes{} 
    \item[] Justification: The results include z-score, confidence intervals, p-values, and chi-squared test.  

\item {\bf Experiments compute resources}
    \item[] Question: For each experiment, does the paper provide sufficient information on the computer resources (type of compute workers, memory, time of execution) needed to reproduce the experiments?
    \item[] Answer: \answerYes{} 
    \item[] Justification: Our compute usage was minimal through API calls to the respective models.
    
\item {\bf Code of ethics}
    \item[] Question: Does the research conducted in the paper conform, in every respect, with the NeurIPS Code of Ethics \url{https://neurips.cc/public/EthicsGuidelines}?
    \item[] Answer: \answerYes{} 
    \item[] Justification: Author(s) reviewed the NeurIPS Code of Ethics to ensure our paper is in alignment.

\item {\bf Broader impacts}
    \item[] Question: Does the paper discuss both potential positive societal impacts and negative societal impacts of the work performed?
    \item[] Answer: \answerYes{} 
    \item[] Justification: The paper discusses potential negative societal impacts that can arise when LLMs are used as intended but give truth-bias skewed results.
    
\item {\bf Safeguards}
    \item[] Question: Does the paper describe safeguards that have been put in place for responsible release of data or models that have a high risk for misuse (e.g., pretrained language models, image generators, or scraped datasets)?
    \item[] Answer: \answerNA{} 
    \item[] Justification: The paper poses no such risks.

\item {\bf Licenses for existing assets}
    \item[] Question: Are the creators or original owners of assets (e.g., code, data, models), used in the paper, properly credited and are the license and terms of use explicitly mentioned and properly respected?
    \item[] Answer: \answerYes{} 
    \item[] Justification: Creators of original data sets are properly credited and the license is explicitly mentioned. 

\item {\bf New assets}
    \item[] Question: Are new assets introduced in the paper well documented and is the documentation provided alongside the assets?
    \item[] Answer: \answerNA{} 
    \item[] Justification: The paper does not release new assets.

\item {\bf Crowdsourcing and research with human subjects}
    \item[] Question: For crowdsourcing experiments and research with human subjects, does the paper include the full text of instructions given to participants and screenshots, if applicable, as well as details about compensation (if any)? 
    \item[] Answer: \answerNA{} 
    \item[] Justification: The paper does not involve crowdsourcing nor research with human subjects.

\item {\bf Institutional review board (IRB) approvals or equivalent for research with human subjects}
    \item[] Question: Does the paper describe potential risks incurred by study participants, whether such risks were disclosed to the subjects, and whether Institutional Review Board (IRB) approvals (or an equivalent approval/review based on the requirements of your country or institution) were obtained?
    \item[] Answer: \answerNA{} 
    \item[] Justification: The paper does not involve crowdsourcing nor research with human subjects.

\item {\bf Declaration of LLM usage}
    \item[] Question: Does the paper describe the usage of LLMs if it is an important, original, or non-standard component of the core methods in this research? Note that if the LLM is used only for writing, editing, or formatting purposes and does not impact the core methodology, scientific rigorousness, or originality of the research, declaration is not required.
    \item[] Answer: \answerNA{} 
    \item[] Justification: The core method development in this research does not involve LLMs as any important, original, or non-standard components

\end{enumerate}

\end{document}